%% file: main.tex
\documentclass[10pt,twocolumn,letterpaper]{article}

\PassOptionsToPackage{table}{xcolor}
\usepackage[preprint]{cvpr}
\input{preamble}

\definecolor{cvprblue}{rgb}{0.21,0.49,0.74}
\usepackage[pagebackref,breaklinks,colorlinks,allcolors=cvprblue]{hyperref}

\def\confName{CVPR}
\def\confYear{2027}

\title{Policy-intrinsic Demonstration Curation \vspace{-0.2em}}
\author{
Sunjae Jeong\textsuperscript{1,2,*} \qquad 
Geunhyeok Yu\textsuperscript{1,*} \qquad 
Dowoon Lee\textsuperscript{1} \qquad 
Hyoseok Hwang\textsuperscript{1} \\
\textsuperscript{1}~Kyung Hee University \qquad 
\textsuperscript{2}~MOBILTECH CO., LTD. \\
{\tt\small \{sunj, geunhyeok, pauldoun, hyoseok\}@khu.ac.kr}
\vspace{-0.5em}
}

\begin{document}
\maketitle
\iftoggle{cvprfinal}{\begingroup\renewcommand{\thefootnote}{}\footnotetext{\textsuperscript{*}Equal contribution.}\endgroup}{}

\begin{abstract}
Fine-tuning a robot policy for a new target task requires target demonstrations, but only a few are available.
Demonstrations of other tasks can fill this gap, and demonstration curation selects the useful ones among these candidates.
Retrieval-based curation selects the candidates that are most similar to the target demonstrations.
Since it does not use the policy, any candidate that looks similar receives a high score.
Performance-aware curation therefore estimates the effect of each candidate on the policy.
However, this requires training on the candidates and takes a long time.
We propose a third branch, policy-intrinsic curation, which uses the policy and still takes little time.
It estimates this effect with forward passes of the pretrained policy and needs no training.
We instantiate this branch with PICKER (Policy-Intrinsic Curation from Known Errors).
It selects the candidates on which the policy makes errors that agree with those on the target demonstrations.
PICKER reaches success rates comparable to those of performance-aware curation and takes about as long as retrieval-based curation.
Its selection is also stable across seeds.
\end{abstract}

\section{Introduction}
\label{sec:intro}

\begin{figure}[t]
  \centering
  \includegraphics[width=\linewidth]{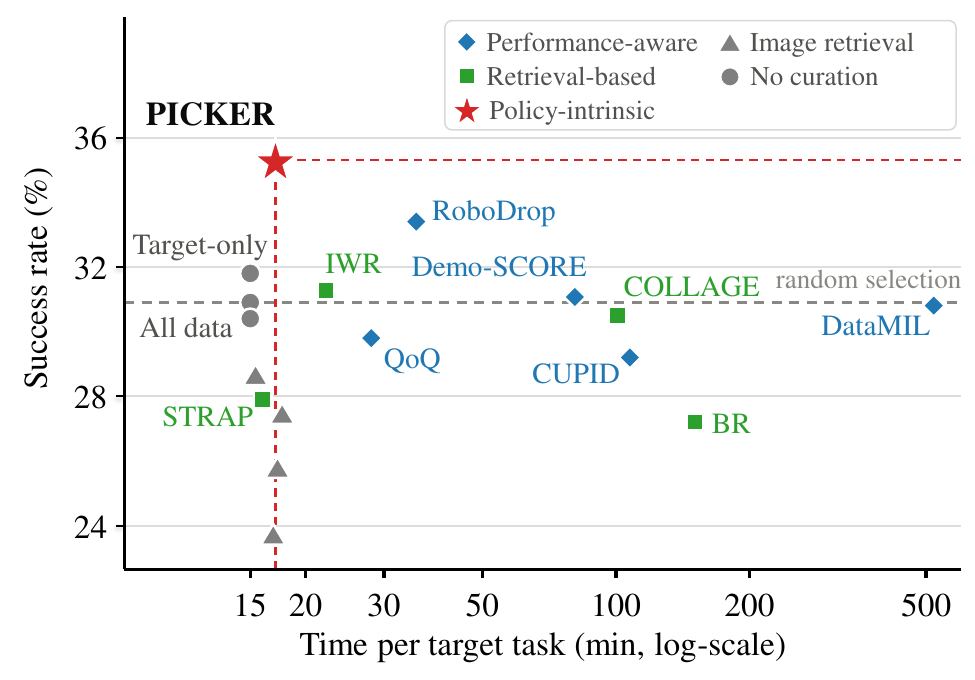}
  \caption{\textbf{Cost vs.\ success rate} on LIBERO-100 with Octo-Small. Displays total time per target task and success rate.}  \vspace{-1.5em}
  \label{fig:tradeoff}
\end{figure}

Robot policies are pretrained on large and diverse collections of demonstrations~\cite{oxe,droid,bridge} and then fine-tuned for new target tasks~\cite{policy-octo,openvla,pi0,groot}.
Fine-tuning requires demonstrations of each target task, which we call target demonstrations.
Each of them is recorded by a human operator who guides the robot, so only a few are available.
Far more demonstrations have already been recorded for other tasks, and adding these candidates to the target demonstrations can fill this gap.
However, Figure~\ref{fig:tradeoff} shows that adding all of them can even lower the success rate, and so can adding a random subset.
This is because only some candidates are useful for the target tasks, while others are unrelated to them or even conflict with them.
Demonstration curation therefore selects the useful candidates, and its result depends on how their usefulness is measured.

\begin{figure*}[t]
  \centering
  \begin{subfigure}[b]{0.1226\textwidth}
    \centering
    \includegraphics[width=\linewidth, page=1, trim=93 32 93 8, clip]{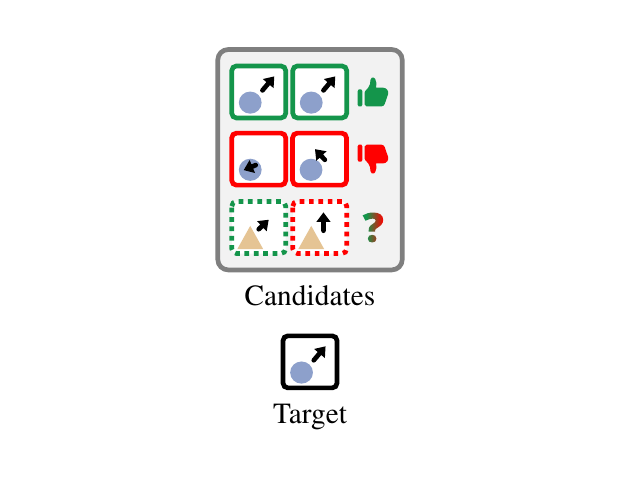}
    \caption{Candidate pool}
    \label{fig:branches-a}
  \end{subfigure}\hfill
  \begin{subfigure}[b]{0.2708\textwidth}
    \centering
    \includegraphics[width=\linewidth, page=2, trim=25 32 26 8, clip]{figures/related_works/fig2-branches.pdf}
    \caption{Retrieval-based}
    \label{fig:branches-b}
  \end{subfigure}\hfill
  \begin{subfigure}[b]{0.2708\textwidth}
    \centering
    \includegraphics[width=\linewidth, page=3, trim=25 32 26 8, clip]{figures/related_works/fig2-branches.pdf}
    \caption{Performance-aware}
    \label{fig:branches-c}
  \end{subfigure}\hfill
  \begin{subfigure}[b]{0.2708\textwidth}
    \centering
    \includegraphics[width=\linewidth, page=4, trim=25 32 26 8, clip]{figures/related_works/fig2-branches.pdf}
    \caption{Policy-intrinsic (ours)}
    \label{fig:branches-d}
  \end{subfigure}
  \caption{\textbf{Three branches of demonstration curation.}
  (a)~Candidates with the action of the target are useful (green), and the others are harmful (red).
  Circles come from the scene of the target and triangles from other scenes, and the bars show their scores.
  (b)~Retrieval-based methods score similarity, so a harmful candidate from the same scene ranks above a useful one from another scene.
  (c)~Performance-aware methods score the effect of a candidate on the policy, but they need training and backward passes.
  (d)~Policy-intrinsic curation (ours) scores this effect with forward passes of the pretrained policy alone.}
  \vspace{-1.5em}
  \label{fig:branches}
\end{figure*}

Existing methods measure this usefulness in one of two ways (Figure~\ref{fig:branches}), and the basic approach is to measure it by similarity.
Retrieval-based methods~\cite{baseline-br,baseline-fr,baseline-strap} select the candidates that are most similar to the target demonstrations in the feature space of a separate model.
Their premise is that a candidate that resembles the target demonstrations is also useful for the target tasks.
However, this similarity does not depend on the policy, so a similar candidate receives a high score even when training on it harms the policy.
For example, a candidate from the scene of a target task looks similar even when its action differs from that of the target demonstrations.
Training on such a candidate can move the prediction of the policy away from the target action.
Similarity alone is therefore not enough to determine whether a candidate helps the policy.

Performance-aware methods~\cite{baseline-qoq,baseline-datamil,baseline-robodrop} instead score a candidate by its effect on the policy, that is, by how training on the candidate would change the policy.
They estimate this effect with tools from data attribution~\cite{koh,tracin,datamodels}, which relate the behavior of a trained model to its training data.
This estimate, however, is itself obtained by training.
A policy is typically first trained on the candidates, and a backward pass is then run for every candidate to obtain its gradient.
Curation thus becomes a training run of its own, which takes a long time.
Its selection also changes with the random seed of this run.
In effect, these methods train a policy to determine where training will lead it.

We start from the observation that this need not be determined by training.
On every demonstration, the pretrained policy makes an error that points from its predicted action to the demonstrated action.
To first order, a training step moves the prediction along this error, so the error shows in advance where training will lead the policy.
On the target demonstrations, these errors show what the policy has yet to learn, and we call them the target errors.
The errors on a candidate can therefore be compared with the target errors.
The comparison indicates whether this candidate would move the policy toward what it has yet to learn.
All of these errors are obtained from forward passes, in which the pretrained policy is evaluated but not updated.

This observation leads to a third branch of curation, which we call policy-intrinsic curation because all of its evidence comes from the pretrained policy itself.
A policy-intrinsic method scores a candidate by its effect on the policy, as performance-aware methods do.
Unlike them, it estimates this effect from forward passes of the pretrained policy alone.
It trains no model and computes no gradient, so curation takes little time.
Policy-intrinsic curation thus combines the strengths of both existing branches (Figure~\ref{fig:tradeoff}).

We instantiate this branch with PICKER.
We call a candidate useful for a target demonstration when the policy represents their states similarly and makes errors in the same direction on both.
The first condition matters because training on a candidate changes the prediction mainly in states that the policy represents similarly.
However, it is not obvious that forward passes suffice, because the effect of one demonstration on another is given by their gradient alignment.
This alignment is the inner product of their gradients.
We show that, at the layer that reads the observation, it separates into one term in the observation representations and one in the errors.
PICKER estimates both terms with forward passes and selects the candidates whose errors agree best with the target errors.

Our contributions are as follows:
\begin{itemize}
  \item We propose policy-intrinsic curation, a third branch of demonstration curation that estimates the effect of each candidate on the policy from forward passes alone.
  \item We present PICKER, which derives a forward-only score from gradient alignment and selects the candidates whose errors agree with the target errors.
  \item On three benchmarks, PICKER reaches the success rates of performance-aware curation at the cost of retrieval-based curation. Its selection is stable across seeds.
\end{itemize}

\section{Related Work}
\label{sec:related}

\begin{figure*}[t]
    \includegraphics[width=\linewidth]{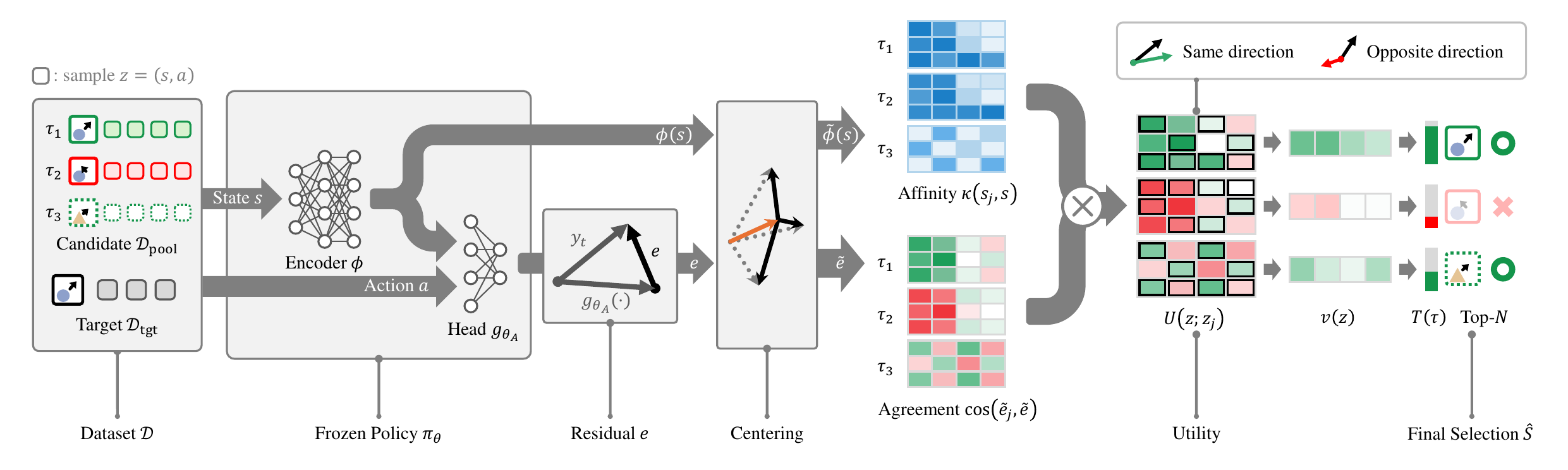}
    \caption{\textbf{Overview of PICKER} for the candidates of Figure~\ref{fig:branches}.
    The pretrained policy processes every candidate sample and target sample with forward passes only.
    Its encoder returns the observation representation $\phi(s)$, and its action head returns the error $e$, which the figure labels as the residual.
    After centering, the affinity compares the representations of a candidate sample and a target sample, and the agreement compares the directions of their errors.
    Their product is the utility $U$, which is averaged into a score $T(\tau)$ for each candidate. PICKER selects the $N$ candidates with the highest scores.}
    \vspace{-1.5em}
    \label{fig:overview}
\end{figure*}

\paragraph{Retrieval-based curation.}
This branch comes from few-shot imitation learning, where a few target demonstrations are augmented with similar data from a large prior dataset (Figure~\ref{fig:branches-b}).
Behavior Retrieval~\cite{baseline-br} compares learned embeddings of states and actions, and FlowRetrieval~\cite{baseline-fr} compares the motion in the demonstrations.
STRAP~\cite{baseline-strap} matches segments of demonstrations with the features of a pretrained vision model, and COLLAGE~\cite{baseline-collage} weights several such measures with reference policies.
DemInf~\cite{deminf} instead scores quality by the mutual information between states and actions.
These methods find relevant or high-quality candidates without estimating the effect of training.
Their scores, however, are computed in a feature space that is separate from the policy.
As a result, they reflect what a candidate shows, not how training on it changes the policy.

\paragraph{Performance-aware curation.}
How training on a candidate changes the policy is the question of data attribution (influence functions~\cite{koh}, gradient-based estimators~\cite{tracin, trak}, and datamodels~\cite{datamodels}).
Performance-aware methods bring these tools to robot demonstrations (Figure~\ref{fig:branches-c}).
QoQ~\cite{baseline-qoq} trains a policy on the pool and compares the gradients of candidates and target demonstrations.
DataMIL~\cite{baseline-datamil} fits datamodels that predict a metric on the target tasks, and RoboDrop~\cite{baseline-robodrop} compares gradients during a warm-up epoch.
CUPID~\cite{baseline-cupid} and Demo-SCORE~\cite{demoscore} instead rely on rollouts, in which the policy is run in the environment.
These methods measure what curation needs, namely the effect of a candidate on the policy.
However, they obtain this effect only after training on the candidates or running the policy, so curation itself requires a training run.

\paragraph{Policy-intrinsic curation.}
Whether this training run is needed depends on what a policy provides before it is trained on the candidates.
Work on data attribution and data selection has approached this question from two sides.
LESS~\cite{less} computes gradients after a short warm-up, and influence scores of the last layer reduce to the similarity of activations multiplied by loss terms~\cite{firstlast}.
Other scores use the error of a model directly, through its norm in data pruning~\cite{el2n} or through its size in interactive imitation learning~\cite{compat}.
These works shorten the training but do not remove it, or they use the error only through its size, without regard to the target task.
Policy-intrinsic curation instead removes the training altogether (Figure~\ref{fig:branches-d}).
PICKER scores each candidate by the direction of its errors relative to the target errors, all obtained from forward passes of the pretrained policy.

\section{Preliminaries}
\label{sec:prelim}

\subsection{Generative Action Policies}
\label{sec:policies}

PICKER takes these errors from the action head of a generative policy.
Demonstrated actions are often multimodal, so many recent policies generate actions with denoising diffusion~\cite{ddpm} or flow matching~\cite{flow-matching}.
A sample $z=(s,a)$ pairs a state $s$ of recent observations with the action chunk $a$ that follows.
Training a policy $\pi_\theta$ on a set of demonstrations $\mathcal{D}$ minimizes
\begin{equation}
  \mathbb{E}_{(s,a)\sim\mathcal{D},\,t,\,\epsilon}\, \tfrac{1}{2}\big\|y_t-g_\theta(x_t,s,t)\big\|^2 ,
  \label{eq:objective}
\end{equation}
where its action head $g_\theta$ takes $s$, a noise level $t$, and a noisy chunk $x_t$ that mixes $a$ with Gaussian noise $\epsilon$.
The head predicts a regression target $y_t$, which is the noise $\epsilon$ for diffusion and the velocity $\epsilon-a$ for flow matching.

\subsection{Demonstration Curation}
\label{sec:setting}

The policy that minimizes Equation~\ref{eq:objective} depends on the demonstrations in $\mathcal{D}$.
Only a small set $\mathcal{D}_{\mathrm{tgt}}$ of successful target demonstrations is available, together with a large pool $\mathcal{D}_{\mathrm{pool}}$ of candidates that vary in task and quality.
For a subset $S\subseteq\mathcal{D}_{\mathrm{pool}}$, fine-tuning the pretrained parameters $\theta_A$ on $\mathcal{D}_{\mathrm{tgt}}\cup S$ yields $\theta(S)$.
Let $J(\pi)$ denote the success rate of a policy $\pi$ on the target tasks.
Given a budget of $N$ demonstrations, the goal of curation is
\begin{equation}
  \max_{S\subseteq\mathcal{D}_{\mathrm{pool}},\ |S|=N}\ J\big(\pi_{\theta(S)}\big).
  \label{eq:selection}
\end{equation}

Computing $\theta(S)$ requires fine-tuning, and evaluating $J$ requires rollouts, so Equation~\ref{eq:selection} cannot be solved directly.
Curation methods therefore rank the candidates with a surrogate score.
We require that this score uses only forward passes of the pretrained policy $\pi_{\theta_A}$.
We train no model, compute no gradient, and run no rollout.
We call such a score policy-intrinsic.

\section{Method}
\label{sec:method}
 
\subsection{Error Alignment}
\label{sec:align}

Figure~\ref{fig:overview} gives an overview of PICKER, which builds such a score on one criterion.
A candidate sample is useful for a target sample $z_j=(s_j,a_j)$ from $\mathcal{D}_{\mathrm{tgt}}$ when two conditions hold.
The pretrained policy represents the two states similarly, and it makes errors in the same direction on both.

To see why, consider a linear action head $\hat a(s)=W\phi(s)$ on a representation $\phi(s)$.
The error on a sample $z=(s,a)$ is $e(z)=a-\hat a(s)$.
To first order, a gradient step on $z$ reduces the squared error $\ell$ of $z_j$ in proportion to the inner product of their gradients.
Since $\nabla_W\ell(z)=-e(z)\,\phi(s)^\top$, this gradient alignment splits into three terms,
\begin{equation}
    \langle\nabla_W\ell(z_j),\nabla_W\ell(z)\rangle=\underbrace{\kappa(s_j,s)}_{\text{affinity}}\,\underbrace{\cos(e_j,e)}_{\text{agreement}}\,\underbrace{\|e_j\|\,\|e\|}_{\text{magnitude}},
  \label{eq:factor}
\end{equation}
where $\kappa(s_j,s)=\langle\phi(s_j),\phi(s)\rangle$, and $e$ and $e_j$ are the errors on $z$ and $z_j$.
For two similar states, a positive agreement means that training on $z$ moves the prediction at $s_j$ toward $a_j$.

There are two obstacles to using Equation~\ref{eq:factor} as a score.
Diffusion and flow-matching heads are nonlinear and take noise as input.
The gradients in Equation~\ref{eq:factor} also require a backward pass for every sample, which our setting excludes.

\subsection{Inside the Action Head}
\label{sec:head}

Even in such a head, the decomposition of Equation~\ref{eq:factor} holds exactly at the observation layer, the linear layer whose input is the observation representation $\phi(s)$.
Let $W$ now denote this layer and $e(z)=y_t-g_{\theta_A}(x_t,s,t)$ the error of the head.
Since each sample receives its own noise draw, a gradient step on $z$ follows the gradient of the noise-averaged loss $\ell(z)$ on average.
The input $\phi(s)$ of this layer does not depend on the noise, so $\nabla_W\ell(z)=-\delta(z)\,\phi(s)^\top$.
Here $\delta(z)$ is the error $e(z)$ back-propagated to the output of $W$ and averaged over noise.
The gradient alignment thus has the same form as Equation~\ref{eq:factor},
\begin{equation}
  \langle\nabla_W\ell(z_j),\nabla_W\ell(z)\rangle=\underbrace{\kappa(s_j,s)}_{\text{affinity}}\,\underbrace{\cos(\delta_j,\delta)}_{\text{agreement}}\,\underbrace{\|\delta_j\|\,\|\delta\|}_{\text{magnitude}}.
  \label{eq:layer}
\end{equation}

For a linear head, $\delta=e$, and Equation~\ref{eq:layer} reduces to Equation~\ref{eq:factor}.
The other layers take the noisy action as input, so their gradients do not split in this way.
For a head that attends to observation tokens, the observation layer consists of the key and value projections.
PICKER then approximates $\phi(s)$ by the mean of the token representations.

\subsection{Scoring without Backward Passes}
\label{sec:score}

Obtaining $\delta$ still requires a backward pass for every sample, so PICKER uses the error $e$ in its place.
This replacement needs only a forward pass and is exact only for a linear head.
PICKER also drops the magnitude, so that the size of the errors does not dominate the selection.

Finally, PICKER evaluates all candidate and target samples under the same $B$ noise draws and subtracts their mean from each error and each representation.
PICKER computes the utility from the resulting centered errors $\tilde e_b$ and $\tilde e_{j,b}$ of $z$ and $z_j$ under the $b$-th draw as
\begin{equation}
  U(z;z_j)=\underbrace{\hat\kappa(s_j,s)}_{\text{affinity}}\,\underbrace{\frac{1}{B}\sum_{b=1}^{B}\cos(\tilde e_{j,b},\tilde e_b)}_{\text{agreement}},
  \label{eq:utility}
\end{equation}
where $\hat\kappa(s_j,s)$ is the cosine similarity of the centered representations.
We call this subtraction centering, and under a shared draw it removes the noise that all errors share.
All terms of Equation~\ref{eq:utility} come from forward passes of $\pi_{\theta_A}$, so the score is policy-intrinsic as defined in Section~\ref{sec:setting}.

\subsection{Selecting Demonstrations}
\label{sec:select}

PICKER scores a sample $z$ by its mean utility $v(z)$ over the set $\mathcal{T}_m(z)$ of the $m$ target samples with the highest utility.
Fine-tuning draws samples uniformly from the selected demonstrations, so the first-order effect of a selection is the average effect of its samples.
PICKER therefore scores each demonstration $\tau$ by the mean $T(\tau)$ of $v(z)$ over its samples,
\begin{equation}
  v(z)=\frac{1}{m}\sum_{z_j\in\mathcal{T}_m(z)}U(z;z_j),\quad T(\tau)=\frac{1}{|\tau|}\sum_{z\in\tau}v(z),
  \label{eq:rule}
\end{equation}
and selects the $N$ candidates with the highest $T(\tau)$.
PICKER averages over the $m$ best matches because a sample need not be relevant to more than a few target samples.
An average over all of them would dilute $v(z)$, and the single best match would make it depend on one target sample.
PICKER runs the forward passes over the pool once and shares them across all target tasks.

\section{Experiments}
\label{sec:experiments}

\subsection{Experimental Setup}
\label{ssec:experiments-setup}

\begin{table*}[t]
  \centering
  \scriptsize
  \setlength{\tabcolsep}{5pt}
  \begin{tabular}{@{}l *{15}{r}@{}}
    \toprule
    & \multicolumn{6}{c}{LIBERO-100} & \multicolumn{6}{c}{LIBERO-X} & \multicolumn{3}{c}{RoboCasa365} \\
    \cmidrule(lr){2-7} \cmidrule(lr){8-13} \cmidrule(lr){14-16}
    & \multicolumn{3}{c}{Octo-Small} & \multicolumn{3}{c}{SmolVLA}
    & \multicolumn{3}{c}{Octo-Small} & \multicolumn{3}{c}{SmolVLA}
    & \multicolumn{3}{c}{Octo-Small} \\
    \cmidrule(lr){2-4} \cmidrule(lr){5-7} \cmidrule(lr){8-10} \cmidrule(lr){11-13} \cmidrule(lr){14-16}
    Curation  & SR & Time & Eff. & SR & Time & Eff. & SR & Time & Eff.
              & SR & Time & Eff. & SR & Time & Eff. \\
    \midrule
    \multicolumn{16}{@{}l}{\textbf{No curation}} \\
    All data & 30.4{\tiny$\pm$6.4} & 15.0 & $-$2.00 & 30.3{\tiny$\pm$6.0} & 98.0 & $+$1.47 & 12.1{\tiny$\pm$0.9} & 23.2 & $-$1.29 & 19.6{\tiny$\pm$3.5} & 39.9 & $+$5.86 & 18.2{\tiny$\pm$0.8} & 131.3 & $-$0.18 \\
    Target-only & 31.8{\tiny$\pm$1.1} & 15.0 & $+$3.60 & 25.2{\tiny$\pm$5.0} & 98.0 & $-$1.65 & 16.9{\tiny$\pm$0.5} & 23.2 & $+$11.12 & 15.1{\tiny$\pm$3.3} & 39.9 & $-$0.90 & 17.6{\tiny$\pm$0.6} & 131.3 & $-$0.46 \\
    Random & 30.9{\tiny$\pm$0.8} & 15.0 & -- & 27.9{\tiny$\pm$5.0} & 98.0 & -- & 12.6{\tiny$\pm$1.5} & 23.2 & -- & 15.7{\tiny$\pm$4.1} & 39.9 & -- & 18.6{\tiny$\pm$0.9} & 131.3 & -- \\
    \midrule
    \multicolumn{16}{@{}l}{\textbf{Performance-aware}} \\
    QoQ & 29.8{\tiny$\pm$2.9} & 28.1 & $-$2.35 & 27.6{\tiny$\pm$2.6} & 181.1 & $-$0.10 & 11.9{\tiny$\pm$1.3} & 32.7 & $-$1.28 & 18.5{\tiny$\pm$2.7} & 106.0 & $+$1.58 & 19.0{\tiny$\pm$1.0} & 202.0 & $+$0.12 \\
    DataMIL & 30.8{\tiny$\pm$0.9} & 520.3 & $-$0.01 & 29.1{\tiny$\pm$8.6} & 2087.4 & $+$0.03 & 11.9{\tiny$\pm$1.2} & 652.1 & $-$0.06 & 16.8{\tiny$\pm$2.2} & 2562.2 & $+$0.03 & 18.2{\tiny$\pm$0.4} & 194.4 & $-$0.12 \\
    RoboDrop & 33.4{\tiny$\pm$2.1} & 35.5 & $+$4.23 & 31.3{\tiny$\pm$1.1} & 173.1 & $+$1.18 & 11.4{\tiny$\pm$1.0} & 42.4 & $-$1.70 & 16.7{\tiny$\pm$2.4} & 107.8 & $+$0.56 & 19.5{\tiny$\pm$1.5} & 349.7 & $+$0.15 \\
    \midrule
    \multicolumn{16}{@{}l}{\textbf{Retrieval-based}} \\
    BR & 27.2{\tiny$\pm$3.2} & 150.8 & $-$1.47 & 32.8{\tiny$\pm$4.1} & 233.9 & $+$1.26 & 11.6{\tiny$\pm$0.3} & 114.1 & $-$0.53 & 18.6{\tiny$\pm$2.6} & 130.8 & $+$1.33 & 18.6{\tiny$\pm$0.3} & 227.1 & 0.00 \\
    STRAP & 27.9{\tiny$\pm$2.9} & 16.0 & $-$11.25 & 29.5{\tiny$\pm$4.7} & 99.0 & $+$0.97 & 9.7{\tiny$\pm$0.9} & 24.3 & $-$7.16 & 21.2{\tiny$\pm$4.9} & 41.0 & $+$8.05 & 19.8{\tiny$\pm$0.4} & 151.9 & $+$0.47 \\
    COLLAGE & 30.5{\tiny$\pm$0.7} & 100.8 & $-$0.24 & 28.4{\tiny$\pm$5.1} & 285.8 & $+$0.10 & 10.9{\tiny$\pm$0.6} & 112.9 & $-$0.90 & 19.8{\tiny$\pm$2.5} & 231.6 & $+$1.06 & $^{\dagger}$19.2{\tiny$\pm$0.2} & 993.7 & $+$0.04 \\
    \midrule
    \multicolumn{16}{@{}l}{\textbf{Policy-intrinsic (ours)}} \\
    \rowcolor{gray!15} \textbf{PICKER} & \textbf{35.3}{\tiny$\pm$1.7} & \textbf{17.1} & \textbf{$+$15.44} & \textbf{32.9}{\tiny$\pm$2.7} & \textbf{107.5} & \textbf{$+$2.81} & \textbf{18.0}{\tiny$\pm$0.8} & \textbf{25.2} & \textbf{$+$12.86} & \textbf{24.1}{\tiny$\pm$3.5} & \textbf{48.5} & \textbf{$+$10.39} & \textbf{19.5}{\tiny$\pm$0.9} & \textbf{181.3} & \textbf{$+$0.30} \\
    \bottomrule
  \end{tabular}
  \caption{\textbf{Main results.}
  SR (\%) of the final policy with its standard deviation over training seeds, total time per target task (min), and efficiency (Eff.).
  $\dagger$ marks point clouds that are reconstructed with VGGT~\cite{vggt}.}
  \vspace{-1.5em}
  \label{tab:main}
\end{table*}

\begin{table}[t]
  \centering
  \footnotesize
  \renewcommand{\arraystretch}{0.92}
  \begin{tabular}{@{}lrrr@{}}
    \toprule
    Method & Task-shared & Per task & Amortized per task \\
    \midrule
    QoQ      & 129.3  & 0.2   & 13.1  \\
    DataMIL  & --     & 505.3 & 505.3 \\
    RoboDrop & 194.6  & 1.1   & 20.5  \\
    BR       & 1356.4 & 0.2   & 135.8 \\
    STRAP    & 7.1    & 0.3   & 1.0   \\
    COLLAGE  & 214.9  & 64.3  & 85.8  \\
    \midrule
    \rowcolor{gray!15} \textbf{PICKER} & \textbf{19.7} & \textbf{0.2}   & \textbf{2.1}   \\
    \bottomrule
  \end{tabular}
  \caption{\textbf{Breakdown of the curation time (min) on LIBERO-100 with Octo-Small.}
  Table~\ref{tab:main} uses the amortized time, which adds one tenth of the task-shared time to the per-task time.}
  \vspace{-1.8em}
  \label{tab:time}
\end{table}

\paragraph{Datasets.}
We use four robot datasets.
LIBERO-100~\cite{dset-libero-100} is our main benchmark, with the 4,500 demonstrations of LIBERO-90 as the pool and the ten tasks of LIBERO-10 as target tasks.
Its pool shares scenes and objects with the target tasks, so useful and misleading candidates are mixed.
LIBERO-X~\cite{dset-libero-x} has a pool that resembles the target tasks less, with 15 target instructions and 2,114 candidates.
RoboCasa365~\cite{dset-robocasa365} tests curation at scale, with 17 atomic target tasks and 29,732 candidates.
The robomimic dataset~\cite{dset-robomimic} labels the demonstrations of each task by operator proficiency, which lets us measure the quality of a selection.

\paragraph{Policies.}
We fine-tune two pretrained policies with different kinds of action heads, Octo-Small~\cite{policy-octo} with a DDPM~\cite{ddpm} diffusion head and SmolVLA~\cite{policy-smolvla} with a flow-matching~\cite{flow-matching} action expert.
PICKER computes its score with their public checkpoints, without any update.

\paragraph{Baselines.}
We compare PICKER with methods of both existing branches.
The retrieval-based methods are Behavior Retrieval (BR)~\cite{baseline-br}, STRAP~\cite{baseline-strap}, and COLLAGE~\cite{baseline-collage}.
The performance-aware methods are DataMIL~\cite{baseline-datamil}, QoQ~\cite{baseline-qoq}, and RoboDrop~\cite{baseline-robodrop}.
Three variants without curation use all candidates, the target demonstrations alone, or a random subset of the same size.
COLLAGE requires point clouds, which we reconstruct with VGGT~\cite{vggt} on RoboCasa365 (dagger in Table~\ref{tab:main}).
The appendix details how we adapt the official implementations of the baselines.

\paragraph{Protocol.}
In all experiments, PICKER evaluates every second step of each candidate under $B=4$ noise draws and averages the utility over the $m=64$ best target samples.
In Table~\ref{tab:main}, the budget $N$ is 450 demonstrations on LIBERO-100 and LIBERO-X and 2,973 on RoboCasa365.
Fine-tuning takes 20k steps, or 300k on RoboCasa365, with the target demonstrations filling half of each batch.
On LIBERO-100 with Octo-Small, each target task has five target demonstrations.
Each method returns one selection for all ten tasks, and one policy is fine-tuned on it.
We repeat fine-tuning with three seeds and evaluate each policy with 150 rollouts per task.

\paragraph{Metrics.}
We report the success rate (SR) of the final policy, averaged over the target tasks, and state its differences in percentage points.
The total time is the curation time per target task plus the time to fine-tune and evaluate the final policy.
All times are measured in minutes on one RTX 3090.
The curation time includes the stages that all target tasks share, amortized over these tasks.
Efficiency is the gain in SR over random selection per GPU-hour of total time.

\subsection{Main Results}
\label{ssec:experiments-main-results}

Table~\ref{tab:main} tests our claim that PICKER attains the SR of performance-aware curation and takes about as long as retrieval-based curation.
On LIBERO-100 and LIBERO-X, PICKER attains the highest mean SR with both policies, by 0.1 to 2.9 points.
On RoboCasa365, the SR lies between 17.6\% and 19.8\% for all methods, and PICKER is 0.3 points below the best one.
PICKER is the only method that improves over all three variants without curation in all five settings of Table~\ref{tab:main}.
It also matches or exceeds every performance-aware method in each setting, although it never trains on the candidates.

\begin{figure*}[t]
  \centering
  \includegraphics[width=\textwidth]{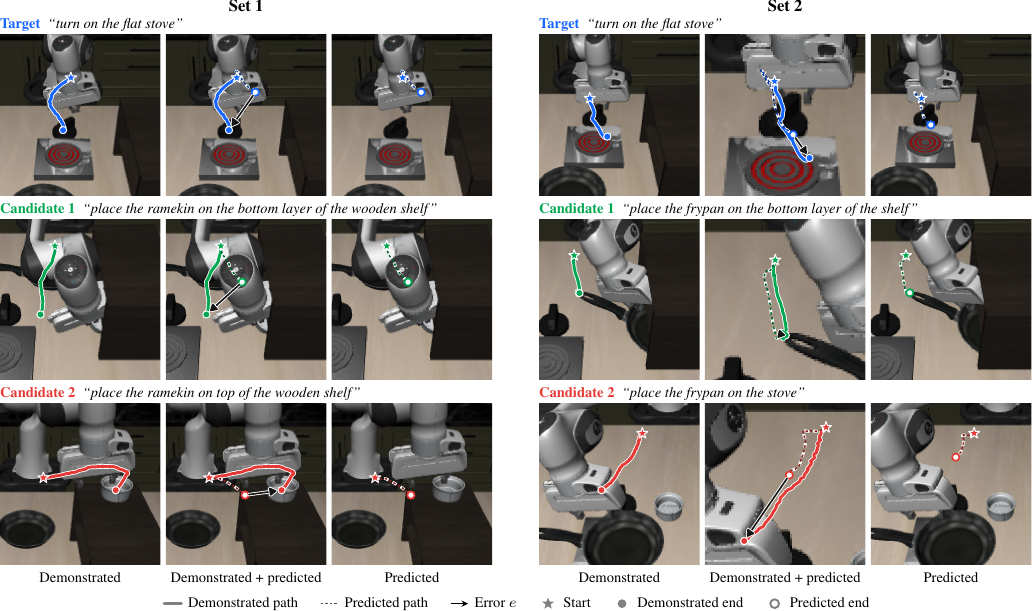}
  \caption{\textbf{Similarity is not usefulness.}
  Two sets from LIBERO-X with SmolVLA, each with one sample of a target demonstration and of two candidates.
  Each row shows the demonstrated and the predicted path of the end effector over the 50 steps of a sample.
  The arrow shows the error at the end, projected onto the image.
  In Set~2, the middle panels are magnified twice.}
  \vspace{-1.5em}
  \label{fig:qualitative}
\end{figure*}

PICKER attains these success rates in little time, because its curation consists of forward passes alone.
Table~\ref{tab:time} divides the curation time into a task-shared stage, which we amortize over the ten target tasks, and a per-task stage.
PICKER is among the fastest methods in both stages, with 19.7 minutes for one forward pass over the pool and 0.2 minutes per task.
Its curation therefore takes 2.1 minutes per task, against 13.1 to 505.3 minutes for the performance-aware methods.
For a single target task, without amortization, it takes 19.9 minutes.
BR trains an encoder and COLLAGE trains reference policies, so STRAP is the only curation method that is faster than PICKER.

Its total time therefore stays within ten minutes of that of random selection in all four LIBERO settings.
Since PICKER also gains the most SR there, it has the highest efficiency in these settings, up to 3.7 times that of the second-best method.
On the larger pool of RoboCasa365, its curation takes about 850 minutes, or 50.0 minutes per target task.
Its efficiency there is second to that of STRAP.

\subsection{Qualitative Analysis}
\label{ssec:experiments-qualitative}

These results rest on one premise, namely that the errors of the pretrained policy show the effect of a candidate before any training.
To examine this premise, we trace single decisions back to the two terms of the score.
Figure~\ref{fig:qualitative} shows two sets from LIBERO-X with SmolVLA, each with one sample of a target demonstration and one sample of each of two candidates.
The predicted path accumulates the actions that the pretrained policy estimates from the first observation and the noisy demonstrated actions.
Table~\ref{tab:sample_selection} lists the affinity and the agreement of these samples, and whether each method selects the candidates within the budget of 450.
The gripper contributes more than half of the agreement in all four cases, so the paths show only a small part of it.

\begin{table}[t]
  \centering
  \footnotesize
  \setlength{\tabcolsep}{4.5pt}
  \newcommand{\qsel}[1]{\textcolor{green!60!black}{$\checkmark$}\,{\tiny\textcolor{gray}{(#1)}}}
  \newcommand{\qrej}[1]{\textcolor{red}{$\times$}\,{\tiny\textcolor{gray}{(#1)}}}
  \newcommand{\qpos}[1]{\textcolor{green!60!black}{\textbf{#1}}}
  \newcommand{\qneg}[1]{\textcolor{red}{{\boldmath$-$}\textbf{#1}}}
  \newcommand{\qpart}[1]{{\scriptsize #1}}
  \begin{tabular}{@{}lcccc@{}}
    \toprule
    & \multicolumn{2}{c}{Set 1} & \multicolumn{2}{c}{Set 2} \\
    \cmidrule(lr){2-3}\cmidrule(l){4-5}
    & Cand.~1 & Cand.~2 & Cand.~1 & Cand.~2 \\
    \midrule
    \multicolumn{5}{@{}l}{\textbf{Selection of the candidate (rank)}} \\
    Affinity-only & \qrej{612}  & \qsel{279}  & \qsel{3}   & \qsel{40}   \\
    STRAP         & \qrej{1505} & \qrej{1441} & \qsel{110} & \qsel{418}  \\
    DataMIL       & \qrej{1638} & \qrej{1806} & \qsel{272} & \qrej{972}  \\
    QoQ           & \qsel{31}   & \qrej{1434} & \qsel{17}  & \qrej{1652} \\
    \rowcolor{gray!15} \textbf{PICKER} & \qsel{309} & \qrej{913} & \qsel{3} & \qrej{568} \\
    \midrule
    \multicolumn{5}{@{}l}{\textbf{Displayed sample}} \\
    Affinity  & 0.543 & 0.722 & 0.703 & 0.760 \\
    Agreement & \qpos{0.635} & \qneg{0.320} & \qpos{0.478} & \qneg{0.234} \\
    \qpart{\quad Position} & \qpart{0.100} & \qpart{$-$0.072} & \qpart{0.028} & \qpart{0.024} \\
    \qpart{\quad Rotation} & \qpart{0.175} & \qpart{$-$0.043} & \qpart{0.079} & \qpart{$-$0.076} \\
    \qpart{\quad Gripper}  & \qpart{0.360} & \qpart{$-$0.205} & \qpart{0.371} & \qpart{$-$0.181} \\
    Target loss decrease (\%) & \qpos{17.58} & \qneg{12.05} & \qpos{6.55} & \qneg{4.43} \\
    \bottomrule
  \end{tabular}
  \caption{\textbf{The two sets of Figure~\ref{fig:qualitative}.}
  A check marks a candidate that a method selects with the budget of 450.
  The gray number is its rank among the 2,114 candidates.
  The smaller rows divide the agreement of the displayed samples by action component.
  The target loss is that of the other target demonstrations after one training step on the sample.}
  \vspace{-1.8em}
  \label{tab:sample_selection}
\end{table}

PICKER multiplies the two terms and averages the product over the samples of a candidate.
A high score therefore requires samples that combine a high affinity with a positive agreement.
The two sets differ in what the affinity alone selects.
In Set~1, the sample of Candidate~2 has the higher affinity, and a selection by affinity alone selects this candidate and excludes Candidate~1.
In Set~2, the two samples have similar affinities, and this selection includes both candidates.

The agreement, in contrast, separates the two candidates in both sets.
It is negative for Candidate~2, which PICKER ranks outside the budget.
It is positive for Candidate~1, which PICKER selects in both sets.
STRAP compares visual features, and it selects neither candidate in Set~1 and both in Set~2.
QoQ selects and excludes the same candidates as PICKER, but it must first train a policy on the pool.

We then test directly whether the agreement indicates the effect of a candidate on the target.
We take one training step of equal norm on each candidate sample and measure the loss on the other target demonstrations before and after.
A step on the sample of Candidate~1 lowers this loss in both sets, and a step on that of Candidate~2 raises it.
In these four cases, a higher agreement corresponds to a larger decrease of the loss.
The affinity does not show this effect, because it is higher for the candidate whose sample raises the loss.

In both sets, PICKER therefore selects the candidate whose sample lowers the loss on the target, and not the one with the higher affinity.
Neither STRAP nor the affinity alone selects this candidate without the one whose sample raises the loss.
Forward passes thus suffice to predict the direction of this effect in all four cases.

\subsection{Selection Analysis}
\label{ssec:experiments-selection}

The two sets show single decisions, so we next examine the whole selection.
PICKER computes its score from the errors of the policy, so we expect its selection to follow what a demonstration does, not how it looks.
A pool does not always consist of expert demonstrations, so we first test with Octo-Small whether PICKER selects demonstrations of high quality.
In each of three robomimic tasks, operators of different proficiency demonstrate the same task.
Of the 45 demonstrations that PICKER selects per task, 76\% are labeled ``Better'' and 10\% ``Worse'', against 30\% and 38\% for random selection (Table~\ref{tab:robomimic}).
STRAP selects 63\% ``Better'' demonstrations, and its SR is 0.9 points below that of PICKER.
PICKER thus selects demonstrations of high quality, although its score uses no proficiency label.

\begin{table}[t]
  \centering
  \small
  \setlength{\tabcolsep}{4pt}
  \begin{tabular}{@{}lcccc@{}}
    \toprule
    & Better (\%) & Okay (\%) & Worse (\%) & SR (\%) \\
    \midrule
    Target-only & -- & -- & -- & 35.3{\tiny$\pm$3.1} \\
    Random      & 30 & 33 & 38 & 40.9{\tiny$\pm$1.9} \\
    \midrule
    STRAP          & 63 & 24 & 13 & 57.3{\tiny$\pm$1.2} \\
    w/o agreement  & 59 & 26 & 16 & 52.9{\tiny$\pm$2.5} \\
    \rowcolor{gray!15} \textbf{PICKER} & \textbf{76} & \textbf{14} & \textbf{10} & \textbf{58.2}{\tiny$\pm$2.1} \\
    \bottomrule
  \end{tabular}
  \caption{\textbf{Selection quality on robomimic with Octo-Small.}
  Share of each level of operator proficiency in the selections of 45 demonstrations per task and the resulting SR for Lift, Can, and Square.
  Each task has its own policy and its own pool.
  Without the agreement, the utility is the affinity alone.}
  \vspace{-1.8em}
  \label{tab:robomimic}
\end{table}

We then compare the tasks of the selected demonstrations on LIBERO-100, with 45 demonstrations selected for each target task alone.
PICKER and STRAP share only 15.6 of these 45 demonstrations on average.
The difference is clearest where the scene and the action disagree.
For the target task of placing a book in the back compartment of a caddy, STRAP selects 23 demonstrations of the same scene.
They place the book in the front compartment.
PICKER selects four of them and 22 demonstrations from another scene that place the book in the back compartment.

The second example is the target task of putting a mug in the microwave and closing it.
There, PICKER selects 15 demonstrations that close a drawer in other kitchens.
These demonstrations share the closing action with the target task and not its scene.
Both selections follow what a demonstration does and not how it looks, in line with our expectation.

\subsection{Ablation Study}
\label{ssec:experiments-ablation}

We next test on LIBERO-100 with Octo-Small which components of the score the SR depends on (Table~\ref{tab:ablation}).
The mean SR drops whichever component of Equation~\ref{eq:utility} is removed.
Without the affinity, the utility also counts pairs of samples whose states the policy represents differently, and the SR drops by 3.1 points.
Without the centering, the errors keep the part that all samples share under the same noise draw, and the SR drops by 1.5 points.
The drop is smallest without the agreement, with 1.0 point, which is within one standard deviation.
Read alone, this order gives the impression that the agreement is hardly needed.

\begin{table}[t]
  \centering
  \small
  \setlength{\tabcolsep}{2.5pt}
  \begin{tabular}{@{}lccccccr@{}}
    \toprule
    & Cent. & $\kappa$ & Agr. & $\delta$ proxy & Mag. & SR (\%) & Drop \\
    \midrule
    w/o agreement & \checkmark & \checkmark & & \checkmark & & 34.3{\tiny$\pm$2.7} & 1.0 \\
    w/o affinity  & \checkmark & & \checkmark & \checkmark & & 32.2{\tiny$\pm$1.7} & 3.1 \\
    w/o centering & & \checkmark & \checkmark & \checkmark & & 33.8{\tiny$\pm$3.8} & 1.5 \\
    \midrule
    Exact $\delta$ & \checkmark & \checkmark & \checkmark & & & 31.3{\tiny$\pm$1.9} & 4.0 \\
    Exact Equation~\ref{eq:layer} & n/a & \checkmark & \checkmark & & \checkmark & 32.0{\tiny$\pm$3.3} & 3.3 \\
    \midrule
    \rowcolor{gray!15} \textbf{PICKER} & \checkmark & \checkmark & \checkmark & \checkmark & & \textbf{35.3}{\tiny$\pm$1.7} & -- \\
    \bottomrule
  \end{tabular}
  \caption{\textbf{Ablation of the components on LIBERO-100 with Octo-Small.}
  The columns mark the use of the centering (Cent.), the affinity ($\kappa$), and the agreement (Agr.).
  The $\delta$ proxy uses $e$ in place of $\delta$, and Mag. keeps the magnitude.
  Drop is the decrease in SR from PICKER.
  On robomimic, the drop without the agreement is 5.3 points (Table~\ref{tab:robomimic}).}
  \vspace{-1.8em}
  \label{tab:ablation}
\end{table}

However, the drop without the agreement is not small on every pool.
Without the agreement, PICKER selects by affinity alone, which is still policy-intrinsic.
The candidates of LIBERO-100 differ in task and scene, whereas those of robomimic show the same task and differ in the proficiency of their operators.
On robomimic, removing the agreement lowers the share of ``Better'' demonstrations from 76\% to 59\% and the SR by 5.3 points (Table~\ref{tab:robomimic}).
Both values then lie below those of STRAP, so the advantage of PICKER over this retrieval-based method comes from the agreement.
Tables~\ref{tab:robomimic} and~\ref{tab:ablation} together thus show that the score needs both terms.
They also suggest that the agreement matters more among candidates of the same task.

Table~\ref{tab:ablation} also tests whether the score loses accuracy because it avoids backward passes.
One variant replaces only the error $e$ by the exact $\delta$ from backward passes, and its SR is 4.0 points lower.
A second variant computes Equation~\ref{eq:layer} exactly, with the magnitude and without the centering, and loses 3.3 points.
The demonstrations that this variant selects lie at the 93rd percentile of the pool in gradient magnitude, against the 64th for PICKER.
None of them comes from the scene of a target task.
In this experiment, backward passes thus do not improve the selection.

\subsection{Robustness Analysis}
\label{ssec:experiments-robustness}

Finally, we examine on LIBERO-100 with Octo-Small how PICKER behaves when the budget and the seed change.

\begin{table}[t]
  \centering
  \small
  \setlength{\tabcolsep}{3pt}
  \begin{tabular}{@{}l rrrr rrrr@{}}
    \toprule
    & \multicolumn{4}{c}{SR (\%)} & \multicolumn{4}{c}{IoU (\%)} \\
    \cmidrule(lr){2-5} \cmidrule(l){6-9}
    Budget & 50 & 100 & 250 & 450 & 50 & 100 & 250 & 450 \\
    \midrule
    Random & 33.0 & 30.3 & 35.0 & 33.1 & 1.0 & 1.4 & 3.0 & 5.0 \\
    QoQ    & 32.9 & 33.2 & 35.7 & 31.7 & 4.2 & 5.9 & 11.1 & 14.4 \\
    STRAP  & 33.1 & 33.8 & 35.9 & 35.8 & 23.1 & 34.1 & 44.9 & 48.9 \\
    \midrule
    \rowcolor{gray!15} \textbf{PICKER} & \textbf{37.8} & \textbf{35.7} & \textbf{34.9} & \textbf{33.6} & -- & -- & -- & -- \\
    \bottomrule
  \end{tabular}
  \caption{\textbf{Budget sweep on LIBERO-100 with Octo-Small.}
  SR (\%) and IoU (\%) with the selection of PICKER.
  The budget is per target task, whereas Table~\ref{tab:main} uses one budget for all tasks.
  Each SR averages three seeds.}
  \vspace{-1.0em}
  \label{tab:budget}
\end{table}

\paragraph{Budget.}
Table~\ref{tab:budget} varies the budget per target task from 50 to 450 and reports the intersection over union (IoU) between each selection and that of PICKER.
A small budget keeps only the candidates that a method ranks highest, so it tests the top of its ranking.
With a budget of 50, the baselines stay at the level of random selection, whereas PICKER improves over it by 4.8 points.
With a budget of 100, it improves by 5.4 points, and with 250 and 450 it lies 1.0 and 2.2 points below STRAP.
As the budget grows, the IoU between STRAP and PICKER rises from 23\% to 49\%.
The SR of PICKER is thus highest at the smallest budget.
This suggests that its score ranks the most useful candidates first.

\begin{figure}[t]
  \centering
  \includegraphics[width=\linewidth]{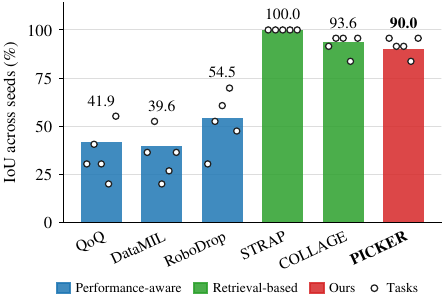}
  \caption{\textbf{Stability across curation seeds on LIBERO-100 with Octo-Small.}
  IoU (\%) between the selections of two seeds.
  Bars average over all tasks, and dots show the five hardest tasks.}
  \vspace{-2.4em}
  \label{fig:seed}
\end{figure}

\paragraph{Seed.}
A curation method should also return the same selection when it is repeated.
The SR does not show this directly, because it also varies with the seed of fine-tuning.
Figure~\ref{fig:seed} therefore compares the selections of two curation seeds by their IoU, with 45 demonstrations per target task.
The selections of the performance-aware methods overlap by only 40\% to 55\%, which we attribute to the model that they train with each seed.
PICKER trains no model, so its only randomness is the noise that it draws for the action head.
Its selections overlap by 90\%, close to the retrieval-based methods.
PICKER thus comes close to the stability of retrieval-based curation while it attains the SR of performance-aware curation.

\section{Conclusion}
\label{sec:conclusion}

We introduced policy-intrinsic curation, a third branch of demonstration curation, and its instance PICKER.
PICKER estimates the effect of each candidate on the policy, as performance-aware methods do, but needs only forward passes of the pretrained policy.
We showed that gradient alignment splits into the affinity between observation representations and the agreement between errors, and PICKER estimates both with forward passes.
On three benchmarks, PICKER reaches success rates comparable to those of performance-aware curation and takes about as long as retrieval-based curation.
On LIBERO-100 and LIBERO-X, it attains the highest mean success rate of all compared methods, and its selection on LIBERO-100 is stable across seeds.
Our analyses indicate that the agreement between errors separates useful candidates from candidates that only look similar.
These results indicate that a pretrained policy already contains the information that is needed to select the demonstrations to learn from.

\paragraph{Limitations.}
PICKER is one instance of policy-intrinsic curation, and its limitations mark directions for further work within this branch.
First, it relies entirely on the pretrained policy.
When this policy makes large errors before fine-tuning, we expect its representations and errors to carry little information about a candidate.
Second, the score is a first-order approximation at the pretrained parameters, which are only the starting point of fine-tuning.
It does not reflect how the usefulness of a candidate changes as the policy is updated.
Third, PICKER runs a forward pass on every candidate, so its curation time grows linearly with the size of the pool.
Its advantage in time therefore becomes smaller on large pools such as RoboCasa365.

\clearpage
{

\section*{LLM Usage Statement}
We used Anthropic Claude to assist with the preparation of this work, including revisions to the manuscript, code, and figures. 
Claude also assisted in designing and conducting experiments, exploring relevant literature, identifying and compiling references, and developing ideas for PICKER. 
We did not use LLMs to generate datasets or annotations. 
All LLM-assisted activities were supervised by the authors, and all resulting materials included in this work were reviewed and verified by the authors.

\small
\bibliographystyle{ieeenat_fullname_unsrt}
\bibliography{main}
}

\end{document}


\raggedbottom
\renewcommand{\thesection}{\Alph{section}}
\renewcommand{\thefigure}{S\arabic{figure}}
\renewcommand{\thetable}{S\arabic{table}}
\renewcommand{\theequation}{S\arabic{equation}}
\renewcommand{\thealgorithm}{S\arabic{algorithm}}

\setcounter{tocdepth}{2}
\twocolumn[{\centering\Large\textbf{\thetitle}\\\vspace{0.5em}Supplementary Material\\\ifshowauthors\vspace{1.0em}{\large\begin{tabular}[t]{c}\paperauthors\end{tabular}}\\\fi\vspace{1.2em}
  \begin{minipage}{0.92\textwidth}\normalsize\raggedright
  {\large\textbf{Table of Contents}}\\[-0.4em]\rule{\linewidth}{0.6pt}
  \vspace{-0.9em}
  \begin{multicols}{2}\supptoc\end{multicols}
  \vspace{-0.9em}
  \rule{\linewidth}{0.6pt}
  \end{minipage}\vspace{1.5em}}]
\ifshowauthors\begingroup\renewcommand{\thefootnote}{}\footnotetext{\textsuperscript{*}Equal contribution.}\endgroup\fi

\noindent\textbf{Overview.}
The main paper refers to this document for the adaptation of the baselines, which Section~\ref{sec:baselines} describes.
The other sections support Section~4 of the main paper.
Section~\ref{sec:derivations} derives the steps from Equation~3 to Equation~5, and Section~\ref{sec:implementation} lists the implementation of PICKER.
Section~\ref{sec:edelta} measures how much the score changes when it uses the exact $\delta$ in place of the error $e$.
Section~\ref{sec:fig1} defines the methods of Figure~1 that Table~1 does not include.
The equations, tables, and figures of this document carry the prefix S.
Numbers without this prefix refer to the main paper.

\section{Derivations}
\label{sec:derivations}

\subsection{Notation}
\label{sec:notation}

This section derives the steps from Equation~3 to Equation~5 of the main paper, and we first fix the notation.
A sample $z=(s,a)$ pairs a state $s$ with an action chunk $a$ in normalized coordinates.
A noise draw $\omega=(t,\epsilon)$ turns this chunk into the noisy chunk $x_t$ and the regression target $y_t$ of Equation~1,
\begin{equation}
  \begin{gathered}
    \underbrace{x_t=\sqrt{\bar\alpha_t}\,a+\sqrt{1-\bar\alpha_t}\,\epsilon,\quad y_t=\epsilon}_{\text{diffusion}},\\
    \underbrace{x_t=(1-t)\,a+t\,\epsilon,\quad y_t=\epsilon-a}_{\text{flow matching}}.
  \end{gathered}
  \label{eq:noisy}
\end{equation}
The head predicts the regression target, and a backward pass carries its error to the output of the observation layer $W$,
\begin{equation}
  \begin{gathered}
    e(z;\omega)=y_t-g_{\theta_A}(x_t,s,t),\qquad h=W\phi(s)+r,\\
    J=\frac{\partial g_{\theta_A}}{\partial h},\qquad \delta(z;\omega)=J^\top e(z;\omega),\\
    \ell(z)=\mathbb{E}_{\omega}\,\tfrac12\big\|e(z;\omega)\big\|^2,\qquad \delta(z)=\mathbb{E}_{\omega}\,\delta(z;\omega).
  \end{gathered}
  \label{eq:notation}
\end{equation}
Here $h$ is the output of the observation layer.
The term $r$ collects the inputs of the head that depend on the noisy chunk and on the noise level.
The Jacobian $J$ belongs to the layers after the observation layer, so $\delta$ is the error back-propagated to the output of this layer.
Without a draw as argument, the loss and the back-propagated error are means over the draws.

\subsection{The Observation Layer}
\label{sec:layer}

We first show that the decomposition of Section~4.2 in the main paper is exact at the observation layer.
The input $\phi(s)$ of this layer does not depend on the draw.
The gradient of the layer under one draw is therefore a rank-one matrix, and so is its mean over the draws,
\begin{equation}
  \begin{gathered}
    \nabla_W\ell(z;\omega)=-\delta(z;\omega)\,\phi(s)^\top,\\
    \nabla_W\ell(z)=-\delta(z)\,\phi(s)^\top,\\
    \big\langle\nabla_W\ell(z_j),\nabla_W\ell(z)\big\rangle=\kappa(s_j,s)\,\langle\delta_j,\delta\rangle,\\
    \mathbb{E}\,\Delta\ell(z_j)=-\eta\,\kappa(s_j,s)\,\langle\delta_j,\delta\rangle+O(\eta^2).
  \end{gathered}
  \label{eq:layersplit}
\end{equation}
The third line holds because the inner product of two rank-one matrices is the product of the inner products of their factors.
It is Equation~4 with the affinity $\kappa$ of Equation~3, before the norms of the two errors are split off.
The fourth line is the expected change of the loss of $z_j$ after a gradient step of size $\eta$ on $z$ in this layer.
The step draws its own noise, and the loss of $z_j$ averages over an independent draw, so both errors are means over the draws.

In Octo-Small, the action head gives one vector to its score network, the mean of the readout tokens.
The score network first applies a linear layer to this vector, together with the noisy chunk and an embedding of the noise level.
The observation layer is the block of this linear layer that multiplies the vector.
The checkpoint has a single readout token, so the representation is one vector.
Equation~4 is thus exact in all experiments with Octo-Small.

\subsection{Several Tokens}
\label{sec:tokens}

This exact decomposition needs one representation vector for each sample.
A head that attends to observation tokens applies the observation layer to each of the $K$ tokens of a sample.
The gradient of this layer is then a sum of $K$ rank-one terms.
We write $\bar\phi$ and $\bar\delta$ for the means over the tokens and $C$ for the part in which the two vary together,
\begin{equation}
  \begin{gathered}
    \nabla_W\ell=-\sum_{k=1}^{K}\delta_k\,\phi_k^\top=-K\big(\bar\delta\,\bar\phi^\top+C\big),\\
    C=\frac1K\sum_{k=1}^{K}(\delta_k-\bar\delta)(\phi_k-\bar\phi)^\top,\\
    \big\langle\nabla_W\ell(z_j),\nabla_W\ell(z)\big\rangle\\
    =K_jK\big(\langle\bar\phi_j,\bar\phi\rangle\,\langle\bar\delta_j,\bar\delta\rangle+R\big),
  \end{gathered}
  \label{eq:tokens}
\end{equation}
where $K_j$ is the number of tokens of $z_j$, and $R$ collects the terms that contain the part $C$ of either sample.
The mean of the token representations keeps the first term.
This term is exact when the error and the representation do not vary together across the tokens of a sample.
The part $C$ then vanishes.

The key and value projections read the same tokens.
Their affinity is therefore shared, and their two errors are concatenated.

\subsection{Errors and Action Errors}
\label{sec:action}

We now turn from the representation to the error.
Let $\hat a$ be the clean chunk that the head predicts in one step from the noisy chunk.
Inserting the noisy chunk of Equation~\ref{eq:noisy} into the error shows that the error is a multiple of the action error,
\begin{equation}
  \begin{gathered}
    \underbrace{\hat a=\frac{x_t-\sqrt{1-\bar\alpha_t}\,g_{\theta_A}}{\sqrt{\bar\alpha_t}},\quad c_t=\sqrt{\frac{\bar\alpha_t}{1-\bar\alpha_t}}}_{\text{diffusion}},\\
    \underbrace{\hat a=x_t-t\,g_{\theta_A},\quad c_t=\frac1t}_{\text{flow matching}},\\
    e(z;\omega)=-c_t\,\big(a-\hat a(z;\omega)\big).
  \end{gathered}
  \label{eq:action}
\end{equation}
Under a shared draw, all samples have the same factor $c_t$, and centering keeps it.
The cosine of two centered errors under one draw is thus the cosine of two centered action errors.
The agreement therefore compares errors in the space of actions, like the error of Equation~3.

The factor $c_t$ also explains why Equation~5 takes the cosine for each draw.
For diffusion, its square is the signal-to-noise ratio of the draw.
In the schedule of Octo-Small, this ratio falls from 124 to $6\times10^{-6}$ over the 20 diffusion steps, which spans more than seven orders of magnitude.
The mean of the errors over the draws, as in Equation~4, weights each draw by $c_t$.
The mean of their inner products weights it by $c_t^2$.
Both means are thus dominated by the draws with the least noise.
The cosine for each draw removes $c_t$ together with the size of the action error, so all $B$ draws receive the same weight.

\subsection{From \texorpdfstring{$\delta$ to $e$}{delta to e}}
\label{sec:delta}

The agreement of Equation~5 uses $e$ in place of $\delta$, which needs no backward pass.
In this section, all errors are those of two samples under one shared draw.
Consider first two samples that also share the Jacobian $J$.
If $JJ^\top$ is a multiple of the identity, the cosine of the two back-propagated errors equals the cosine of the two errors.
For a general Jacobian of full row rank, the two cosines still have the same sign when the errors are aligned well enough,
\begin{equation}
  \begin{gathered}
    |\cos(e_j,e)|>\frac{\kappa_J^2-1}{\kappa_J^2+1},\qquad\kappa_J=\frac{\sigma_{\max}}{\sigma_{\min}}\\
    \Longrightarrow\quad\cos(\delta_j,\delta)\,\cos(e_j,e)>0,
  \end{gathered}
  \label{eq:sign}
\end{equation}
where $\sigma_{\max}$ and $\sigma_{\min}$ are the largest and the smallest singular value of the Jacobian, and $\kappa_J$ is its condition number.

For the proof, let $x$ and $y$ be the two errors scaled to unit norm, with a positive cosine and the angle $\theta$ between them.
The polarization identity and the singular values of the Jacobian give
\begin{equation}
  \begin{aligned}
    4\,\langle J^\top x,J^\top y\rangle&=\big\|J^\top(x+y)\big\|^2-\big\|J^\top(x-y)\big\|^2\\
    &\ge\sigma_{\min}^2\,\|x+y\|^2-\sigma_{\max}^2\,\|x-y\|^2\\
    &=4\cos^2\tfrac{\theta}{2}\,\Big(\sigma_{\min}^2-\sigma_{\max}^2\tan^2\tfrac{\theta}{2}\Big).
  \end{aligned}
  \label{eq:polar}
\end{equation}
The last line is positive when $\tan(\theta/2)<1/\kappa_J$, which is the condition of Equation~\ref{eq:sign}.
The case of a negative cosine follows with $-y$ in place of $y$.
More generally, the condition number bounds how much a linear map changes any angle~\cite{bauer1960}.

When the Jacobians of the two samples differ, the same statement holds up to a small change of angle.
Suppose that both Jacobians lie close to a reference Jacobian $\bar J$ of full row rank, at a relative distance of at most $\rho<1$.
With $\angle$ for the angle between two vectors, this distance bounds the change of angle,
\begin{equation}
  \begin{gathered}
    \max\big(\|J_j-\bar J\|,\,\|J-\bar J\|\big)\le\rho\,\sigma_{\min}(\bar J)\\
    \Longrightarrow\quad\big|\angle(\delta_j,\delta)-\angle(\bar J^\top e_j,\bar J^\top e)\big|\le2\arcsin\rho.
  \end{gathered}
  \label{eq:jacobians}
\end{equation}
The bound holds because each back-propagated error lies within the angle $\arcsin\rho$ of its counterpart under the reference Jacobian.

In Octo-Small, the layers after the observation layer depend on the input only through its output $h$, so the Jacobian is a function of $h$.
Under a shared draw, the embedding of the noise level and the noise cancel in the difference of this output between two samples,
\begin{equation}
  h_j-h=W(\phi_j-\phi)+\sqrt{\bar\alpha_t}\,W_x\,(a_j-a),
  \label{eq:hdiff}
\end{equation}
where $W_x$ is the block of the first layer that multiplies the noisy chunk.
Two samples with similar representations and similar actions therefore have similar Jacobians.

Fine-tuning updates all parameters and not only the observation layer.
To first order, a gradient step on $z$ then changes the loss of $z_j$ through the neural tangent kernel $\Theta$ of the head.
This kernel is a sum of one term per layer,
\begin{equation}
  \begin{gathered}
    \Delta\ell(z_j)=-\eta\,e_j^\top\,\Theta(z_j,z)\,e+O(\eta^2),\\
    \Theta(z_j,z)=\sum_{l}\frac{\partial g_j}{\partial\theta_l}\Big(\frac{\partial g}{\partial\theta_l}\Big)^{\!\top},\\
    \frac{\partial g_j}{\partial W}\Big(\frac{\partial g}{\partial W}\Big)^{\!\top}=\kappa(s_j,s)\,J_jJ^\top,
  \end{gathered}
  \label{eq:ntk}
\end{equation}
where $g_j$ and $g$ are the outputs of the head for the two samples, and $\theta_l$ holds the parameters of layer $l$.
The term of the observation layer gives the alignment of Equation~4.
If the term of every layer is a multiple of the identity, the change is proportional to the inner product of the two errors.
In this case, $e$ is the error that all layers share, whereas $\delta$ belongs to one layer.
Section~\ref{sec:edelta} measures how much the replacement changes the score.

\subsection{Centering of the Representations}
\label{sec:centerphi}

The affinity of Equation~5 uses centered representations, and this section derives this step.
Equation~3 describes one gradient step, whereas fine-tuning takes many steps.
We therefore also consider a linear head with a bias and train it to convergence.
Training starts from the pretrained head $\hat a_0$ and minimizes the squared error on a set $\mathcal{Z}$ of $n$ samples.
This set holds the target samples and the selected samples.

We write $\tilde\phi$ for the representation minus its mean on this set.
We assume that the scatter matrix $\Sigma_{\mathcal Z}$ of $\tilde\phi$ on this set is invertible.
For every state $s'$ and every target sample $z_j$, the trained head $\hat a_\infty$ and its error $e_\infty$ then satisfy
\begin{equation}
  \begin{gathered}
    \hat a_\infty(s')-\hat a_0(s')=\sum_{i\in\mathcal Z}H(s',s_i)\,e_0(z_i),\\
    H(s',s_i)=\frac1n+\tilde\phi(s')^\top\,\Sigma_{\mathcal Z}^{-1}\,\tilde\phi(s_i),\\
    e_\infty(z_j)=e_0(z_j)-\sum_{i\in\mathcal Z}H(s_j,s_i)\,e_0(z_i),
  \end{gathered}
  \label{eq:converged}
\end{equation}
where $e_0$ is the error of the pretrained head.

For the proof, let the rows of $X$ hold a one followed by the representation of each sample, and let $Y$ hold the actions.
Gradient descent from the pretrained parameters $\beta_0$ converges to the least squares solution
\begin{equation}
  \beta_\infty=\beta_0+(X^\top X)^{-1}X^\top(Y-X\beta_0),
  \label{eq:lsq}
\end{equation}
and the block inverse of $X^\top X$ gives the weights $H$.

Equation~\ref{eq:converged} has the structure of Equation~3.
Each sample of the set subtracts its error, weighted by $H$, from the error of the target sample.
To first order, the squared error of the target sample thus falls by twice this weight times the inner product of the two errors.
Apart from the constant $1/n$, the weight is an inner product of centered and whitened representations.
The errors are those before training, which forward passes of the pretrained policy provide.

PICKER uses the cosine of the centered representations, because the scatter matrix depends on the selection.
The set $\mathcal{Z}$ is also unknown before the selection, so PICKER centers on all candidate and target samples.

Second-order estimates of influence use the same affinity.
K-FAC approximates the curvature of a layer with bias by a Kronecker product~\cite{kfac}.
Its input factor $A$ is the second moment of the input extended by a one.
With the mean $\mu$ and the covariance $\Sigma$ of the input, the block inverse of this factor gives
\begin{equation}
  \begin{bmatrix}\phi_j\\1\end{bmatrix}^{\!\top}\!A^{-1}\begin{bmatrix}\phi\\1\end{bmatrix}=1+(\phi_j-\mu)^\top\,\Sigma^{-1}\,(\phi-\mu).
  \label{eq:kfac}
\end{equation}
Influence functions for diffusion models build on this approximation~\cite{diffinfluence}.
Centering both the inputs and the errors of a layer is also an established remedy for poorly conditioned gradients~\cite{lecun1998,schraudolph1998}.

\subsection{Centering of the Errors}
\label{sec:centere}

Equation~5 also centers the errors under each draw.
We fix one draw and omit its index.
All samples then receive the same noise $\epsilon$, and centering subtracts the mean over the centering set.
We write $g_j$ for the output of the head on $z_j$ and mark means over the centering set with a bar.
The centered errors are then
\begin{equation}
  \begin{gathered}
    \underbrace{\tilde e_j=-(g_j-\bar g)}_{\text{diffusion}},\\
    \underbrace{\tilde e_j=-(a_j-\bar a)-(g_j-\bar g)}_{\text{flow matching}},
  \end{gathered}
  \label{eq:centered}
\end{equation}
so the shared noise cancels exactly for both heads.

This cancellation holds for $e$ but not for $\delta$.
When the Jacobian differs across samples, centering and the backward pass do not commute.
For a diffusion head, the two orders give
\begin{equation}
  \begin{gathered}
    \underbrace{J_j^\top\,\tilde e_j=-J_j^\top\,(g_j-\bar g)}_{\text{centering first}},\\
    \underbrace{\delta_j-\bar\delta=(J_j-\bar J)^\top\epsilon-\big(J_j^\top g_j-\overline{J^\top g}\big)}_{\text{centering afterwards}}.
  \end{gathered}
  \label{eq:orders}
\end{equation}
In the first order, the mean of the back-propagated errors over the centering set need not vanish.
These errors then keep a part that all samples share.
In the second order, the shared noise remains through the differences between the Jacobians.
In either order, $\delta$ lacks the exact cancellation of the centered $e$.
This is consistent with Table~5 of the main paper, where the exact $\delta$ does not improve the selection.

\subsection{Shared and Stratified Draws}
\label{sec:draws}

This cancellation rests on draws that all samples share.
The $B$ draws act as common random numbers for all candidate and target samples.
The differences between candidates then carry no independent noise, and centering removes the response that all samples share.

The draws are also stratified.
Each draw takes its noise level from one of $B$ ranges of equal probability.
Let $\bar X$ be the mean of a quantity $X$, such as a cosine, over these draws.
The law of total variance gives
\begin{equation}
  \operatorname{Var}(\bar X)=\frac{1}{B^2}\sum_{b=1}^{B}\operatorname{Var}(X\mid b)\le\frac{1}{B}\operatorname{Var}(X).
  \label{eq:strata}
\end{equation}
The right-hand side is the variance for $B$ independent draws from the full range.
The stratification thus removes the variance between the ranges.
The draws depend on the curation seed, which is the only randomness of PICKER (Figure~5 of the main paper).

\subsection{Sign and Top-\texorpdfstring{$m$}{m}}
\label{sec:remarks}

Two remarks concern the sign of the utility and the mean over the $m$ best matches.

\paragraph{Sign.}
For a linear head, a gradient step of size $\eta$ on $z$ changes the prediction at $s_j$ and the squared error of $z_j$ by
\begin{equation}
  \begin{gathered}
    \Delta\hat a(s_j)=\eta\,\kappa(s_j,s)\,e(z),\\
    \Delta\|e_j\|^2=-2\eta\,\kappa(s_j,s)\,\langle e_j,e\rangle+O(\eta^2).
  \end{gathered}
  \label{eq:signrule}
\end{equation}
The error of the target sample falls whenever the affinity and the inner product of the two errors have the same sign.
This includes a negative affinity together with a negative agreement.
In this case, the step moves the prediction at the target state against the error of the candidate and thus toward the target action.
PICKER therefore does not clip the affinity.

\paragraph{Top-\texorpdfstring{$m$}{m}.}
The mean utility over all target samples would depend on the share of target samples of each task.
Suppose that $z$ has the utility $u$ with the $n_k$ target samples of task $k$ and none with the other target samples.
Its mean utility over all $n$ target samples is then
\begin{equation}
  \frac1n\sum_{j=1}^{n}U(z;z_j)=\frac{n_k}{n}\,u,
  \label{eq:topm}
\end{equation}
which grows with the length of the target demonstrations of this task.
The mean over the $m$ best matches does not change when target samples of lower utility are added.
It removes this dependence whenever $m$ is at most the number of target samples of a task.
This holds for $m=64$ in all our settings.

\section{Implementation of PICKER}
\label{sec:implementation}

This section lists how PICKER computes its score in all experiments, and Algorithm~\ref{alg:picker} summarizes the steps.
The forward passes run once over the pool and are shared by all target tasks.

\begin{algorithm}[t]
  \caption{PICKER}
  \label{alg:picker}
  \small\raggedright
  \begin{algorithmic}[1]
    \REQUIRE pretrained policy $\pi_{\theta_A}$, pool $\mathcal{D}_{\mathrm{pool}}$, \mbox{target demonstrations $\mathcal{D}_{\mathrm{tgt}}$}, budget $N$
    \STATE $\mathcal{C}\gets$ samples at every second step of $\mathcal{D}_{\mathrm{pool}}$ \COMMENT{candidates}
    \STATE $\mathcal{T}\gets$ samples at every step of $\mathcal{D}_{\mathrm{tgt}}$ \COMMENT{targets}
    \STATE Draw $t_b\sim p_t(\cdot\mid R_b)$ and $\epsilon_b\sim\mathcal{N}(0,I)$ for $b\le B$
    \FORALL[forward passes only]{$z=(s,a)\in\mathcal{C}\cup\mathcal{T}$}
      \STATE $\phi(s)$ and $e_b(z)\gets y_{t_b}-g_{\theta_A}(x_{t_b},s,t_b)$ for $b\le B$
    \ENDFOR
    \STATE Subtract from $\phi$ and each $e_b$ its mean over $\mathcal{C}\cup\mathcal{T}$
    \FORALL{$z=(s,a)\in\mathcal{C}$}
      \STATE $U(z;z_j)\gets\hat\kappa(s_j,s)\,\frac{1}{B}\sum_b\cos(\tilde e_{j,b},\tilde e_b)$ \COMMENT{Eq.~5}
      \STATE $v(z)\gets$ mean of $U(z;z_j)$ over the $m$ best $z_j\in\mathcal{T}$
    \ENDFOR
    \STATE $T(\tau)\gets$ mean of $v(z)$ over the samples of $\tau$ in $\mathcal{C}$
    \RETURN the $N$ candidates with the highest $T(\tau)$
  \end{algorithmic}
\end{algorithm}

\paragraph{Samples and draws.}
For a selection for several target tasks, $\mathcal{T}$ holds the target samples of all these tasks.
We use $B=4$ draws and the $m=64$ best matches, and the draws depend only on the curation seed.
Each draw takes its noise level from the range $R_b$ and uses one noise vector $\epsilon_b$ for the whole action chunk.
For Octo-Small, $p_t$ is uniform over the 20 diffusion steps, and $R_b$ holds the steps $5b-5$ to $5b-1$.
For SmolVLA, $p_t$ is $\mathrm{Beta}(1.5,1)$ over the flow time, and $R_1,\dots,R_4$ are its four ranges of equal probability.

\paragraph{Representations and errors.}
For Octo-Small, the representation is the readout token at the last step of the observation window, a vector of 384 entries (Section~\ref{sec:layer}).
Its error is the error of the predicted noise on a chunk of four steps, each with seven action dimensions.
For SmolVLA, the representation is the mean of the token representations of the vision-language model, as in Section~4.2 of the main paper.
Its error is the error of the predicted velocity on a chunk of 50 steps with the same seven action dimensions.
Both errors are computed on normalized actions.

\paragraph{Checkpoints.}
PICKER computes its score with the public checkpoints, without any update.
These are \texttt{rail-berkeley/octo-small-1.5} for Octo-Small and \texttt{lerobot/smolvla\_base} at revision \texttt{5e8d12a} for SmolVLA.
Both are the base checkpoints that their authors released, and neither has been fine-tuned on any of our datasets.
SmolVLA was pretrained on community datasets of the SO100 robot~\cite{policy-smolvla}.

\section{Agreement of \texorpdfstring{$e$ and $\delta$}{e and delta}}
\label{sec:edelta}

Section~4.3 of the main paper replaces $\delta$ by $e$, and Table~5 shows that the exact $\delta$ does not raise the SR.
We now measure how much this replacement changes the score itself.
On LIBERO-100 with Octo-Small, we run Algorithm~\ref{alg:picker} once with $e$ and once with the exact $\delta$ of Table~5 in place of $e$.
A superscript marks the error that a quantity uses, and $A$ is the agreement of Equation~5, so that $U=\hat\kappa A$.
Table~\ref{tab:edelta} compares the two runs from the demonstration scores down to the pairs that enter the score.
The two selections share 409 of their 450 demonstrations, so the replacement keeps most decisions of the selection.

\begin{table}[t]
  \centering
  \small
  \setlength{\tabcolsep}{5pt}
  \renewcommand{\arraystretch}{1.2}
  \begin{tabular}{@{}llrr@{}}
    \toprule
    Level & Measure & $n$ & Value \\
    \midrule
    Demonstrations & $\rho_s(T^{e},T^{\delta})$ & 4,500 & 0.99 \\
    Selections & $|S^{e}\cap S^{\delta}|$ & 450 & 409 \\
    Candidate samples & $\rho_s(v^{e},v^{\delta})$ & 335,614 & 0.96 \\
    Pairs in the score & $\Pr[A^{e}A^{\delta}>0]$ & 128,000 & 99.8\% \\
    & $\rho_s(U^{e},U^{\delta})$ & 128,000 & 0.88 \\
    \bottomrule
  \end{tabular}
  \caption{\textbf{Score with $e$ and with the exact $\delta$} on LIBERO-100 with Octo-Small.
  The exact $\delta$ is that of Table~5 of the main paper, and $\rho_s$ is the Spearman correlation over the $n$ items of a level.
  The pairs take 2,000 random candidate samples, each with its $m=64$ best target samples.}
  \vspace{-1em}
  \label{tab:edelta}
\end{table}

\section{Baselines}
\label{sec:baselines}

This section describes how we adapt the baselines of Table~1 (Section~5.1 of the main paper).
All methods select whole candidates, and each method returns one selection of $N$ candidates for the target tasks of a setting.
The final policy is then fine-tuned on this selection and the target demonstrations in the same way for all methods.
We therefore use each method only for its selection.
We do not use its own way of training the policy, such as weighting the selected data.

\paragraph{QoQ.}
QoQ~\cite{baseline-qoq} scores a candidate sample by its highest gradient similarity to the target samples.
The gradients are taken at the action head of a policy that is trained on the pool, and a random projection compresses them~\cite{oporp}.
A candidate receives the mean score of its samples.

\paragraph{DataMIL.}
DataMIL~\cite{baseline-datamil} estimates how each candidate sample changes the loss on the target demonstrations.
It obtains this effect from metagradients, which are derivatives through the training of the policy~\cite{mgd}.
We rank the candidates by this effect, averaged over the samples of each candidate.

\paragraph{RoboDrop.}
RoboDrop~\cite{baseline-robodrop} scores a candidate sample by how well the gradient of its action head aligns with the gradients of target samples that look similar.
It computes these gradients at several checkpoints during the training of a policy on the pool and averages the resulting scores.

\paragraph{BR.}
Behavior Retrieval~\cite{baseline-br} trains a variational autoencoder on the samples of the pool.
It retrieves the samples whose latent lies close to that of a target sample.
We rank the candidates by their share of retrieved samples.

\paragraph{STRAP.}
STRAP~\cite{baseline-strap} matches sub-trajectories of the target demonstrations to the pool with subsequence dynamic time warping on DINOv2 features~\cite{dinov2}.
We rank the candidates by the number of matches that they contain.

\paragraph{COLLAGE.}
COLLAGE~\cite{baseline-collage} retrieves candidates separately with visual, motion, shape, and language features.
It trains a reference policy on the retrievals of each feature type and measures how well this policy explains the target demonstrations.
It then divides the budget among the feature types according to this measure.

\section{Methods of Figure~1}
\label{sec:fig1}

Figure~1 of the main paper also shows four methods that Table~1 does not include, all on LIBERO-100 with Octo-Small.
Each of them selects 45 candidates for each target task, and we merge these selections into one selection of 450 candidates.
We then fine-tune the final policy on this selection and the target demonstrations.

\paragraph{IWR.}
IWR~\cite{iwr} weights the candidates by the ratio between the density of the target data and that of the pool.
We estimate both densities with kernel density estimates.
The estimates use the features of the pretrained Octo-Small, reduced by principal component analysis, together with the action chunks.
A candidate receives the mean log ratio of its samples.

\paragraph{CUPID.}
CUPID~\cite{baseline-cupid} trains a reference policy for each target task and rolls it out.
It scores each candidate by its influence on the success of these rollouts, which it computes with TRAK~\cite{trak} features of the action head.

\paragraph{Demo-SCORE.}
Demo-SCORE~\cite{demoscore} uses the same rollouts to train a classifier that separates successful from failed rollouts.
It scores each candidate with this classifier.

\paragraph{Image retrieval.}
Image retrieval embeds the camera frames with a pretrained image encoder and compares each candidate with the target demonstrations of a task.
Max-sim scores each frame of the candidate by its highest cosine similarity to a target frame and averages over the frames.
Trajectory mean compares the mean embedding of the candidate with that of each target demonstration and keeps the best match.
kNN vote finds the ten nearest candidate frames of each target frame and counts how many of these frames each candidate contains.
Figure~1 shows four configurations of encoder and rule.
ResNet-50~\cite{resnet} uses max-sim, and MAE-B~\cite{mae} uses trajectory mean.
DINOv2-S and DINOv2-B~\cite{dinov2} both use kNN vote.

{
  \footnotesize
  \bibliographystyle{ieeenat_fullname_unsrt}
  \bibliography{main,supplementary}
}

%% file: preamble.tex
\usepackage[T1]{fontenc} 
\usepackage{amsthm,verbatim}
\usepackage{array,multirow}
\usepackage{algorithm}
\usepackage{algorithmic}
\usepackage{microtype}

\usepackage{lipsum}
\usepackage{subcaption}
\usepackage{xcolor}

\theoremstyle{definition}

\theoremstyle{plain}

\theoremstyle{remark}